\documentclass[10pt,conference,letterpaper]{IEEEtran}

\usepackage{amsmath}
\usepackage{array}
\usepackage{booktabs}
\usepackage{cite}
\usepackage{colortbl}
\usepackage{graphicx}
\usepackage{xcolor}
\usepackage{url}
\usepackage{balance}
\definecolor{hdrslate}{HTML}{3B4252}
\definecolor{oldwarm}{HTML}{F5F0EB}
\definecolor{oldwarm2}{HTML}{EDE6DE}
\definecolor{newcool}{HTML}{E8EEF2}
\definecolor{newcool2}{HTML}{DAE3EA}
\definecolor{dimlabel}{HTML}{F7F7F7}
\definecolor{dimlab2}{HTML}{ECECEC}
\definecolor{softgreen}{HTML}{E4EEE6}
\definecolor{softlav}{HTML}{EDE4F0}
\definecolor{softlav2}{HTML}{E3D8E8}

\newcommand{\M}{\mathcal{M}}
\newcommand{\TopK}{\operatorname{TopK}}

\begin{document}

\title{Memory Has Geometry: Non-Uniform Geometric Memory for Long-Horizon Personalized AI}

\author{
\IEEEauthorblockN{Jiahong Liu\textsuperscript{1},
Wenhao Yu\textsuperscript{1},
Zexuan Qiu\textsuperscript{1},
Menglin Yang\textsuperscript{2},
Irwin King\textsuperscript{1}}
\IEEEauthorblockA{\textsuperscript{1}The Chinese University of Hong Kong, Hong Kong SAR, China\\
\{jiahong.liu21, yuwenhao117\}@gmail.com, \{zxqiu22, king\}@cse.cuhk.edu.hk}
\IEEEauthorblockA{\textsuperscript{2}The Hong Kong University of Science and Technology (Guangzhou), Guangzhou, China\\
menglinyang@hkust-gz.edu.cn}
}

\maketitle

\begin{abstract}
Long-term memory is becoming a core substrate for personalized AI, yet most systems still represent personalization as discrete records in a largely static latent space, accessed under one global similarity notion. For data mining, this creates a mismatch: the evidence is a \textbf{temporal event stream}, while the dominant abstraction is a \textbf{searchable record set}. We argue that long-horizon personalization should instead model memory as a \textbf{user-specific dynamical state space} with locally heterogeneous geometry. Geometry here is a computational language, not a literal claim about cognition: it captures stable versus volatile regions, variable-rate drift, heterogeneous neighborhoods, and uncertainty about current user state. Profiles and isolated events remain useful as points, but interaction, feedback, and elapsed time induce trajectories. Memory access then becomes \textbf{trajectory-conditioned reconstruction} of the relevant user state, not only nearest-neighbor lookup.
\end{abstract}


\section{The Memory Bottleneck}

Personalized AI now sustains months-long interactions, but its memory infrastructure has not kept pace. Memory must carry evidence across interactions and maintain an evolving estimate of user state: preferences, constraints, goals, project context, and prior corrections. Dialogue systems \cite{li2025helloagain,ong2025theanine,tan2025prospect,liu2025palace,kim2025share,chen2025ppa}, personalized LLMs \cite{salemi2024lamp,ning2024userllm,zhang2025chameleon,wu2025aligning,liu2025pllm,perfit}, autonomous agents \cite{park2023generative,packer2023memgpt,zhang2024survey}, and recommenders \cite{zhou2018din,zhou2019dien,kang2018sasrec,shani2022surrogate} face the same bottleneck: as histories grow from sessions to lifetimes, personalization is limited less by model capacity than by how experience is stored, structured, and accessed. \textbf{This is a data mining problem at its core}: inferring durable but evolving state from heterogeneous, asynchronous, weakly labeled user-event data.

\begin{figure*}[t]
    \centering
    \includegraphics[width=\textwidth,page=1]{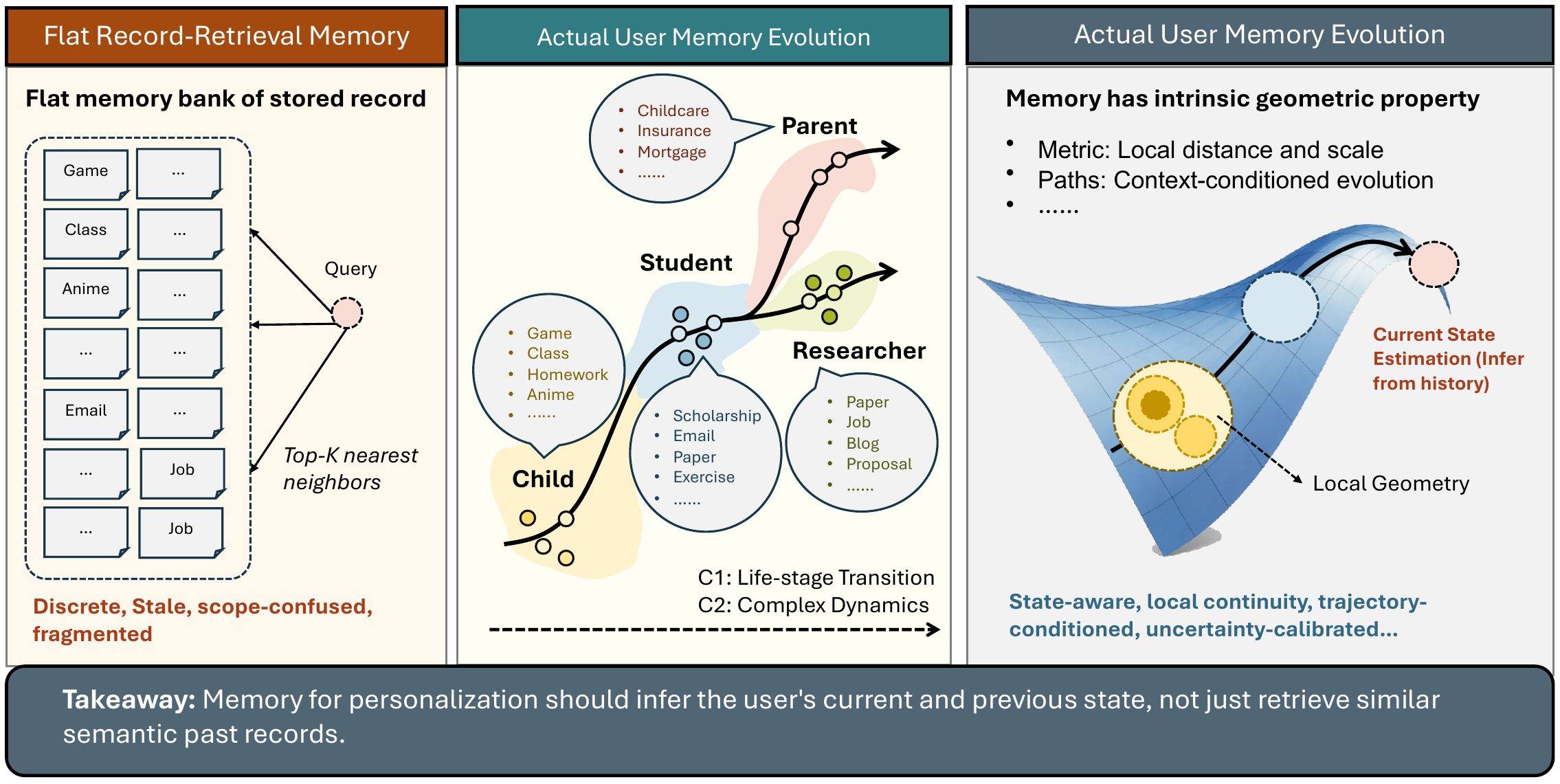}
    \caption{Flat record-retrieval memory versus a geometric view of user memory evolution. Instead of retrieving isolated past records under one similarity notion, geometric memory infers the user's current state from structured trajectories, local geometry, and longer-term evolution.}
    \label{fig:memory-contrast}
\end{figure*}

Existing systems include textual memory banks \cite{park2023generative,zhong2023memorybank,wang2023longmem}, vector retrieval \cite{guu2020realm,lewis2020rag}, and learned external memories \cite{graves2016dnc}. A-Mem dynamically links notes \cite{xu2025amem}, Zep maintains temporal knowledge graphs \cite{rasmussen2025zep}, and HeLa-Mem consolidates associative memories \cite{liu2026helamem}. Our distinction is explicit local metrics and trajectory-conditioned state estimation, rather than structure alone.

\begin{table}[t]
\caption{Record-retrieval memory versus geometric memory.}
\label{tab:contrast}
\centering
\footnotesize
\setlength{\tabcolsep}{2.2pt}
\begin{tabular}{@{}>{\centering\arraybackslash}p{0.15\columnwidth}>{\centering\arraybackslash}p{0.36\columnwidth}>{\centering\arraybackslash}p{0.36\columnwidth}@{}}
\toprule
\rowcolor{hdrslate}
 & \textcolor{white}{\textbf{Record-Retrieval}} & \textcolor{white}{\textbf{Geometric}} \\
\midrule
\cellcolor{dimlabel}\textbf{Store} & \cellcolor{oldwarm}Discrete records & \cellcolor{newcool}Latent state and trajectories \\[2pt]
\cellcolor{dimlabel}\textbf{Measure} & \cellcolor{oldwarm}Single global similarity & \cellcolor{newcool}Global structure plus local uncertainty \\[2pt]
\cellcolor{dimlabel}\textbf{Update} & \cellcolor{oldwarm}Append, overwrite, summarize & \cellcolor{newcool}Path-dependent dynamical update \\[2pt]
\cellcolor{dimlabel}\textbf{Access} & \cellcolor{oldwarm}Top-$k$ neighbors & \cellcolor{newcool}Trajectory-conditioned reconstruction \\
\bottomrule
\end{tabular}
\end{table}

The structural gap is between a temporal, weakly supervised event process and a searchable record set. Later evidence changes the relevance of earlier events; local corrections need not generalize. Figure~\ref{fig:memory-contrast} groups a toy user's records into distinct interest, routine, constraint, and project trajectories. Retrieval can find relevant evidence while still estimating the wrong current state.

This convergence hides three assumptions. \textbf{A1: discrete records.} Memory is a set of messages, slots, profiles, summaries, or embeddings. \textbf{A2: uniform geometry.} These items live in one latent space governed by one broadly sufficient similarity metric \cite{li2020tisasrec,sun2019bert4rec}. \textbf{A3: item-level access.} Memory is accessed by retrieving or attending to stored items, rather than reconstructing user state. These assumptions make stored items primary, rather than evolving state. \textbf{The bottleneck is therefore not retrieval quality alone, but the abstraction itself.}

\section{A Geometric Memory View}

We propose modeling personalized memory as a \textbf{\emph{geometric dynamical system}}: a user-specific state space whose local metric structure, curvature, scale, and uncertainty vary across regions and evolve with interaction. Geometry is a computational abstraction, not a literal cognitive manifold.

Let $(\M,\mathcal{G})$ be a personalized memory state space, where $\M$ is a manifold-like state space and $\mathcal{G}=\{g_x\}_{x\in\M}$ is a field of positive-definite local metrics encoding scale and anisotropy; curvature is derived from metric variation, while uncertainty is modeled separately. A user's memory state at time $t$ is a point $x_t\in\M$. An event $e_t=(o_t,\Delta t_t)$ contains interaction evidence $o_t$ and elapsed time $\Delta t_t$.

Given state $x_{t-1}$ and event $e_t$, memory evolves as
\[
x_t=\operatorname{Exp}_{x_{t-1}}\!\left(g_{x_{t-1}}^{-1} f_\theta(x_{t-1},e_t)\right),
\]
where $f_\theta$ proposes a tangent covector and the inverse metric converts it to a displacement. With an identity metric this is an additive Euclidean update. A posterior over metrics and states represents epistemic uncertainty; update rate is learned separately, as in continuous-time dynamics \cite{chen2018neuralode}. High curvature does not itself imply rapid drift.

The fundamental object is then a trajectory family $\Gamma=\{\gamma_k\}$, with each path tracing a coherent evolution of user state: a preference regime, project, routine, or episode. Record-centric access retrieves top-$k$ neighbors,
\[
R(q_t,D_t)=\TopK_{m_i\in D_t} s(q_t,m_i).
\]
Geometric memory replaces this with trajectory-conditioned reconstruction,
\[
\hat{r}_t=\int h_\theta(q_t,x_t,z)\,p(z\mid q_t,x_t,\Gamma)\,dz,
\]
where $z$ is a relevant path segment and $\hat{r}_t$ may drive response generation, recommendation, action, adaptation, confidence, or abstention. \textbf{The shift is conceptual but consequential}: memory access becomes inference over trajectories, not lookup over items.

\section{Why This Is a BlueSky Agenda}

The ICDM 2026 BlueSky track asks for bold visions that expose gaps, challenge assumptions, and define transformative research directions \cite{icdm2026bluesky}. Geometric memory fits that call by reframing long-horizon personalization as \textbf{state estimation under non-uniform geometry}, not another retrieval refinement. The central idea is to mine latent dynamics, local uncertainty, and trajectory structure from persistent interaction streams instead of building ever larger stores of retrievable records.

\textbf{Why ICDM?} Sequential recommendation and semantic transfer already model evolving behavior \cite{zhou2019dien,koren2009temporal,zhang2026semacdr,li2026genair}. PerFit identifies shared and user-specific representation shifts \cite{perfit}; HRCF and HICF exploit hyperbolic structure in recommendation \cite{yang2022hrcf,yang2022hicf,yang2025hgformer}. Geometric learning and foundation-model agendas motivate matching representation to structure \cite{bronstein2021geometric,yang2022hgl,he2025position,yang2026geometric,liu2026tutorial}. Our contribution combines position-dependent metrics, trajectory-conditioned reconstruction, and controllable updates. JODIE forecasts user--item embedding trajectories \cite{kumar2019jodie}; temporal graph networks maintain event-driven states \cite{rossi2020tgn}. Geometry adds an explicit distance and transport structure governing how updates propagate across contexts; matched identity-metric ablations can isolate this contribution directly.

\textbf{A concrete instantiation.} Hyperbolic memory is one illustrative case. Hyperbolic spaces support coarse-to-fine branching: broad stable tendencies at one scale, and specific interests, projects, or temporary regimes at finer scales. A user's state could evolve through hyperbolic exponential and logarithmic maps while reconstruction reasons over geodesic segments. Other users or domains may require Euclidean, spherical, mixed-curvature, graph-structured, or hybrid spaces; the agenda is to learn the geometry rather than assume one metric.

\begin{figure}[t]
\centering
\includegraphics[width=\columnwidth]{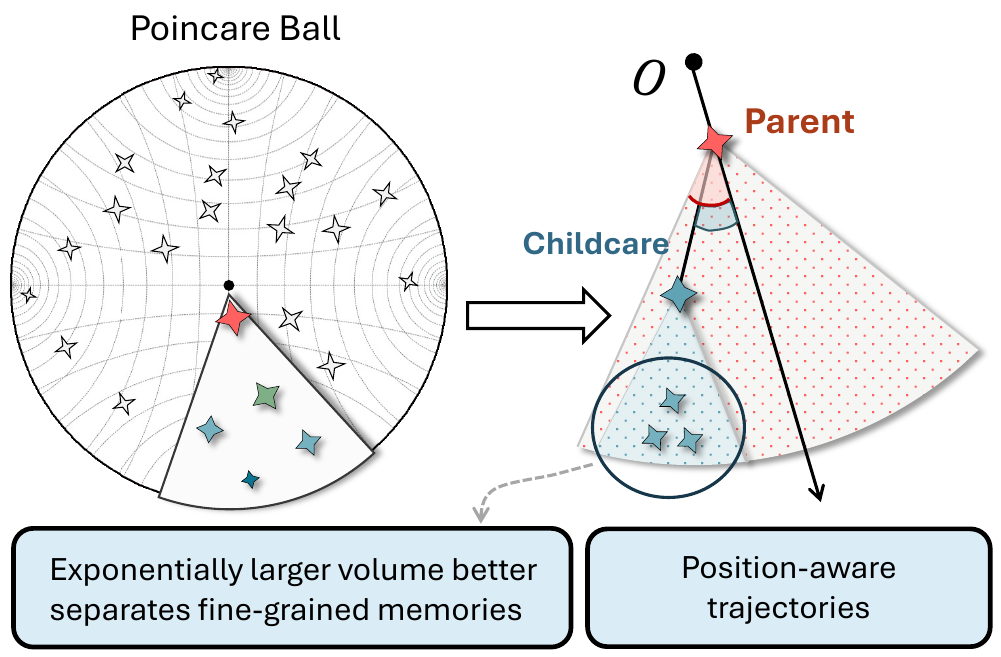}
\caption{Illustrative coarse-to-fine organization of user state, with hyperbolic geometry as one candidate substrate. In a candidate implementation, a broad role anchors context-specific paths, allowing a temporary childcare constraint to remain local rather than overwrite a general preference. Distance measures state compatibility, not elapsed time; arrows indicate inferred transitions. Learned routing and a state decoder complement the layout with temporal dynamics and predictive uncertainty.}
\label{fig:hyperbolic}
\end{figure}

\section{Research Opportunities}

Table~\ref{tab:landscape} summarizes geometric memory research opportunities.

\begin{table*}[t]
\caption{Research landscape of geometric memory.}
\label{tab:landscape}
\centering
\small
\setlength{\tabcolsep}{5pt}
\renewcommand{\arraystretch}{1.07}
\begin{tabular}{@{}>{\centering\arraybackslash}p{0.13\textwidth}>{\raggedright\arraybackslash}p{0.14\textwidth}>{\raggedright\arraybackslash}p{0.62\textwidth}@{}}
\toprule
\rowcolor{hdrslate}
\textcolor{white}{\textbf{Layer}} & \textcolor{white}{\textbf{Direction}} & \textcolor{white}{\textbf{Core Question}} \\
\midrule
\cellcolor{dimlabel}\textbf{Foundation} & \cellcolor{softgreen}Representation & \cellcolor{softgreen}What geometric substrate--Riemannian, mixed-curvature, graph-structured, or hybrid--captures multiscale user state? \\[2pt]
\cellcolor{dimlab2}\textbf{Operation} & \cellcolor{oldwarm}Formation & \cellcolor{oldwarm}How should heterogeneous, asynchronous interactions update the state space and reveal trajectory structure? \\[2pt]
\cellcolor{dimlabel} & \cellcolor{oldwarm2}Access & \cellcolor{oldwarm2}How can systems reconstruct relevant user state by marginalizing over trajectory segments under production latency? \\[2pt]
\cellcolor{dimlab2} & \cellcolor{oldwarm}Evolution & \cellcolor{oldwarm}How does local geometry adapt under preference drift, regime shifts, and multiscale temporal dynamics? \\[2pt]
\cellcolor{dimlabel} & \cellcolor{oldwarm2}Forgetting \& Editing & \cellcolor{oldwarm2}How can memory surgery modify or decay local regions with bounded propagation and user-controllable forgetting? \\[2pt]
\cellcolor{dimlab2}\textbf{System} & \cellcolor{newcool}Cross-User Geometry & \cellcolor{newcool}How can systems align, compare, and transfer knowledge across user-specific geometric structures? \\[2pt]
\cellcolor{dimlabel} & \cellcolor{newcool2}Efficiency & \cellcolor{newcool2}How can expensive operations--exponential maps, parallel transport, curvature estimation--support real-time serving? \\[2pt]
\cellcolor{dimlab2} & \cellcolor{newcool}Trust \& Privacy & \cellcolor{newcool}How can geometric uncertainty, trajectory-level explanation, and privacy guarantees coexist? \\[2pt]
\cellcolor{dimlabel}\textbf{Benchmark} & \cellcolor{softlav}Evaluation & \cellcolor{softlav}What temporal granularity, state-evolution complexity, and evidence sparsity must benchmarks reach? \\[2pt]
\cellcolor{dimlab2} & \cellcolor{softlav2}Geometric Metrics & \cellcolor{softlav2}How should we measure trajectory coherence, predictive calibration, editing locality, and forgetting fidelity? \\
\bottomrule
\end{tabular}
\end{table*}

\subsection{Foundation: Geometric Representation}

The first question is which substrate captures multiscale user state. Hyperbolic embeddings and graph convolutions model hierarchy \cite{nickel2017poincare,sala2018tradeoffs,chami2019hgcn}; spherical components recurring routines; flat subspaces near-linear drift; and graph or mixed-curvature products discrete projects and continuous preferences \cite{gu2019mixed}. A tractable starting point shares an event encoder and transition model across users, then learns small user-specific metric adapters under population priors. A fixed product substrate provides common interfaces while avoiding manifold search. Contrastive learning and UHCone offer weak supervision \cite{liu2022hgcl,yang2024uhcone}; population priors mitigate dimensional collapse \cite{zhang2025collapse}. Predictive validation activates useful local flexibility.

\subsection{Operation: Memory Lifecycle}

\textbf{Formation.} A shared encoder maps each event to content, context, and time features. Soft routing assigns it to active or dormant paths using context compatibility and predictive likelihood. Sustained residuals propose a new branch; repeated contextual matches resume a dormant path; compatible predictive states propose a merge. Posterior routing uncertainty should remain explicit, rather than forcing every event into one trajectory. Branch and merge decisions require validation against annotated episodes and held-out prediction.

\textbf{Access.} Trajectory-conditioned reconstruction asks which path segment explains the current context, not which item is closest. This requires latency-aware marginalization, approximate geodesic search, amortized inference, and confidence estimates that survive long gaps and sparse evidence.

\textbf{Evolution.} User state is non-stationary. Metric scale and transition dynamics should be estimated separately, and restructure only when predictive evidence supports new domains. Short episodes, medium-term projects, and long-term identity evolve at different rates; a useful model should distinguish durable drift from brief excursions, and regime shift from noise. Hyperbolic continual learning supports geometric preservation \cite{liu2026hmcl}, while ProWorld models trajectory progress \cite{liu2026proworld}; both supply components for personalized memory evolution.

\textbf{Forgetting and editing.} Staleness can increase predictive uncertainty and reduce evidence weight; it need not change curvature. Editing intervenes on a local state or path, while deletion must also remove supporting records and invalidate derived states. The key requirement is bounded propagation: changing one neighborhood should not distort unrelated memory.

\subsection{System: Deployment and Trust}

\textbf{Cross-user geometry.} FlatLand demonstrates tailored client geometries with shared aggregation \cite{liu2026flatland}, providing a concrete basis for population priors, cold-start transfer, and efficient user-specific adaptation.

\textbf{Efficiency.} Exponential maps, parallel transport, curvature estimation, and trajectory marginalization are costlier than flat inner products. Riemannian adaptive optimization supports manifold training \cite{becigneul2019adaptive}. Hypformer supplies efficient hyperbolic operators \cite{yang2024hypformer}; HypLoRA and low-rank adaptation provide lightweight adaptation tools \cite{hyplora,yang2025lowrank}; HiHPQ suggests geometry-aware compression \cite{qiu2024hihpq}. Streaming inference still requires validation.

\textbf{Trust and privacy.} Position-dependent uncertainty can tell a system when to abstain; trajectory explanations can show whether a decision follows a stable preference or a transient episode; and privacy guarantees must bound what the latent state reveals. Trustworthy agentic AI also supplies privacy and system-security mechanisms \cite{qi2026trust} that can be layered with geometric uncertainty and trajectory-level auditing.

\subsection{Benchmark: Evaluation for Geometric Memory}

Recent benchmarks probe long-horizon memory--LoCoMo \cite{maharana2024locomo}, LongMemEval \cite{wu2025longmemeval}, and LifeBench \cite{cheng2026lifebench}--but structural gaps remain. First, \textbf{temporal granularity}: user state evolves across hours, weeks, projects, and life stages. Second, \textbf{state-evolution complexity}: evaluations should cover consolidation, partly valid old preferences, and regime shifts. Third, \textbf{evidence implicitness}: weak distributed signals should complement declared facts. Fourth, \textbf{geometry-aware metrics} are missing. We need to measure trajectory coherence, predictive calibration, editing locality, and forgetting fidelity, not only end-task accuracy.

\textbf{Validation protocol.} Construct timestamped streams with recurrent roles, gradual preference drift, a temporary exception, a dormant project that resumes, and a true regime change. Controlled streams with known states and consented annotations establish identifiable behavior before scaling to natural interaction logs. Split chronologically and hold out users, varying time gaps and evidence density. Compare retrieval, a strong dynamic latent-state model, and the same encoder, routing, and transition model with identity versus learned local metrics, matching parameter count, data, and memory budget. Measure state prediction, stale-preference error, cross-role leakage, Brier score, selective risk, edit spillover, and latency; report user-level confidence intervals. TRACE motivates process evaluation beyond final accuracy \cite{chen2026trace}.

\textbf{End-to-end probe.} A user repeatedly prefers concise work emails, asks for one detailed grant proposal, then resumes routine email writing. Context routing should place the proposal on a temporary project path. A feedback-trained metric can make displacement from that path into the general email-preference region costly. Reconstruction should retain concise emails while permitting detailed proposals. A dynamic embedding provides a competitive reference, while the identity-metric ablation quantifies geometry's added value.

\textbf{H1: horizon sensitivity} predicts lower stale-state error as gaps and regime shifts grow. \textbf{H2: calibration} predicts better selective risk at matched coverage. \textbf{H3: sparse evidence} predicts fewer events needed to reach a fixed state-prediction error. Consistent gains over the matched dynamic baseline after controlling for capacity and routing would identify the value contributed specifically by local geometry.

\textbf{What would count as success?} Success means recovering state from weak signals while preserving valid old preferences and keeping temporary exceptions local. After a local correction, unrelated contexts should retain their previous behavior; after deletion, both direct retrieval and derived state should cease to expose the removed evidence. These tests assess controllable revision beyond historical fit.

\section{Feasibility and Safeguards}

\textbf{Structured sharing makes the model learnable and deployable.} A population metric and shared transition model provide a data-efficient starting point, with user adapters activated as evidence accumulates. Canonical coordinates remove equivalent parameterizations, while prediction, calibration, and edit-locality objectives identify useful geometry. Decision-level identifiability accepts coordinate-equivalent models that preserve calibrated predictions and localized edits. Evidence-dependent regularization specializes shared behavior as stable personal structure emerges. Cold-start users inherit the shared model, and identity-metric shrinkage stabilizes adaptation. Low-rank adapters, cached local charts, and approximate trajectory search enable scale; private aggregation, consent-aware deletion, and trajectory audits protect inferred state. These mechanisms turn feasibility and safety into measurable acceptance criteria.

\section{Conclusion}

We introduce non-uniform geometric memory as a unified framework for user states, trajectories, local metrics, uncertainty, and controllable intervention. Trajectory-conditioned estimation allows durable preferences, temporary contexts, and regime changes to coexist. The framework connects temporal data mining, geometric learning, recommendation, and personalized agents, supporting consistent adaptation, interpretable decisions, and targeted editing or forgetting.

Deployment builds directly on existing systems: encoders initialize state, metric adapters specialize geometry, trajectory routers organize context, and reconstruction guides decisions. Modular components use standard feedback, while offline replay and shadow deployment evaluate calibration, latency, and edit locality before rollout. This creates a practical path to enduring personalized intelligence.

\section*{Acknowledgment}

The research presented in this paper was partially supported by the Research Grants Council of the Hong Kong Special Administrative Region, China (CUHK 2300246, RGC C1043-24G), (CUHK 14203425, RGC GRF 2151317).

\newpage
\balance
\bibliographystyle{IEEEtran}
\bibliography{references}

\end{document}